\documentclass[10pt,twocolumn,letterpaper]{article}

\usepackage{cvpr}
\usepackage{times}
\usepackage{epsfig}
\usepackage{graphicx}
\usepackage{amsmath}
\usepackage{amssymb}
\usepackage{siunitx}
\usepackage{adjustbox}
\usepackage{authblk}

\usepackage[pagebackref=true,breaklinks=true,letterpaper=true,colorlinks,bookmarks=false]{hyperref}

\cvprfinalcopy 

\def\cvprPaperID{3826} 

\newcommand{\boldhead}[1]{\vspace{0.03in}\noindent\textbf{#1: }}

\begin{document}

\title{Pixels, voxels, and views: A study of shape representations\\ for single view 3D object shape prediction\vspace{-0.34cm}}

\author[1]{Daeyun Shin}
\author[1]{Charless C. Fowlkes}
\author[2]{Derek Hoiem}
\affil[1]{University of California, Irvine}
\affil[2]{University of Illinois, Urbana-Champaign}
\affil[ ]{\tt\small {\{daeyuns, fowlkes\}@ics.uci.edu\hspace{2em}dhoiem@illinois.edu}}
\affil[ ]{\url{http://www.ics.uci.edu/\~daeyuns/pixels-voxels-views}\vspace{-0.05cm}}

\setlength{\affilsep}{0.7em}
\renewcommand\Authsep{\hspace{1.4em}}
\renewcommand\Authands{\hspace{1.4em}}

\maketitle

\begin{abstract}
The goal of this paper is to compare surface-based and volumetric 3D object shape representations, 
as well as viewer-centered and object-centered reference frames for single-view 3D shape prediction.
We propose a new algorithm for predicting depth maps from multiple
viewpoints, with a single depth or RGB image as input.  By modifying the network and the way 
models are evaluated, we can directly compare the merits of voxels vs. surfaces and 
viewer-centered vs. object-centered for familiar vs. unfamiliar objects, as predicted 
from RGB or depth images.
Among our findings, we show that surface-based methods outperform voxel representations for objects from novel classes and produce higher resolution outputs. 
We also find that using viewer-centered coordinates is 
advantageous for novel objects, while object-centered representations are better for more familiar objects.  
Interestingly, the coordinate frame significantly affects the shape representation learned, 
with object-centered placing more importance on implicitly recognizing the object category 
and viewer-centered producing shape representations with less dependence on category recognition.
\end{abstract}

\section{Introduction}

\begin{figure}[t]
\begin{center}
    \includegraphics[scale=0.15]{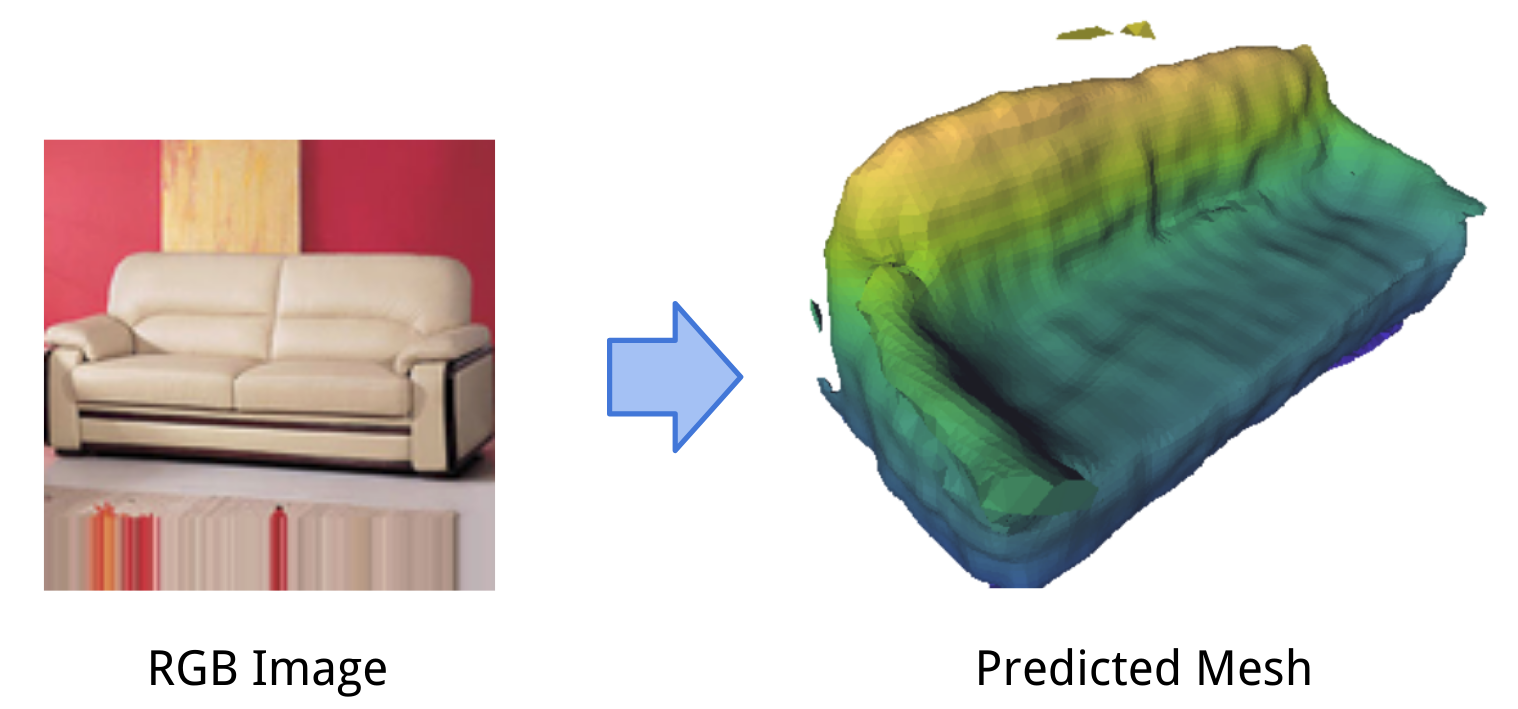}
\end{center}
\vspace{-0.1in}
    \caption{
    We investigate the problem of predicting the 3D shape of an object
    from a single depth or RGB image (illustrated above).
    In particular, we examine the impacts of coordinate frames (viewer-centered vs. object-centered), shape representation (volumetric vs. multi-surface), and familiarity (known instance, novel instance, novel category).
    }
    \vspace{-0.1in}
\label{fig:splash}
\end{figure}

Shape is arguably the most important property of objects, providing cues for affordance, function, category, and interaction. This paper examines the problem of predicting the 3D object shape from a single image (Fig.~\ref{fig:splash}). The availability of large 3D object model datasets~\cite{chang2015shapenet} and flexible deep network learning methods has made this an increasingly active area of research. Recent methods predict complete 3D shape using voxel~\cite{kar2015category,choy20163d} or octree~\cite{tatarchenko2017octree} volumetric representations, multiple depth map surfaces~\cite{tatarchenko2016multi}, point cloud~\cite{fan2017point}, or a set of cuboid part primitives~\cite{zou2017_iccv}.  However, there is not yet a systematic evaluation of important design choices such as the choice of shape representation and coordinate frame.

\begin{figure*}
  \begin{center}
  \adjustbox{trim={.00\width} {.00\height} {0.00\width} {.00\height},clip}{
    \includegraphics[scale=0.1721]{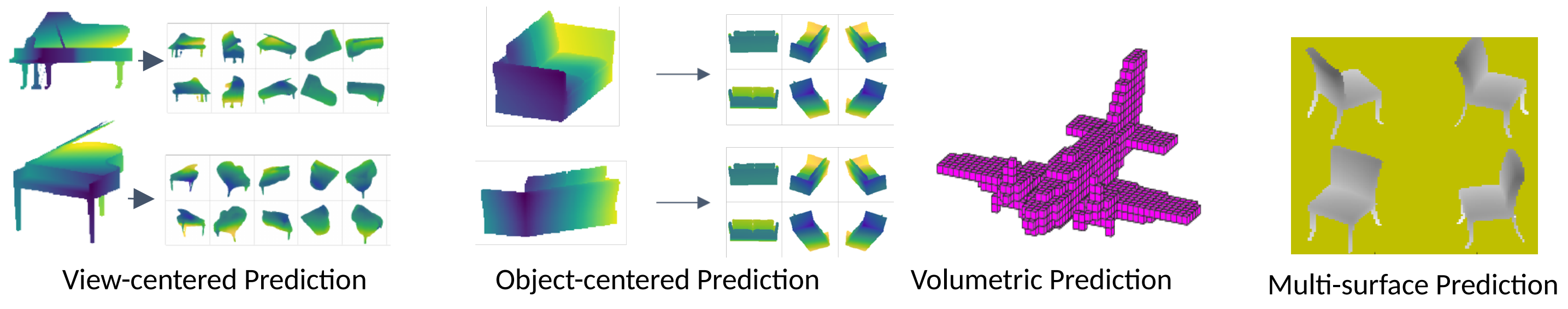}
  }
  \end{center}
  \vspace{-0.1in}
    \caption{We compare view-centered vs. object-centered and volumetric vs. multi-surface formulations of shape prediction. In \textbf{view-centered} prediction, the shape is predicted relative to the viewpoint of the input image, which requires encoding both shape and pose. In \textbf{object-centered} prediction, the shape is predicted in a canonical pose, which is standardized across training and prediction evaluation. For \textbf{volumetric} prediction, the shape is modeled as a set of filled 3D voxels. For \textbf{multi-surface} prediction, the shape is modeled as depth maps from multiple different viewpoints which tile the viewing sphere.}
  \label{fig:terms}
  \vspace{-0.1in}
\end{figure*}

In this paper, we investigate two key issues, illustrated in Fig.~\ref{fig:terms}.  First, is it better to represent shape {\em volumetrically} or as {\em multiple 2.5D surfaces} observed from varying viewpoints? The earliest (albeit still recent) pattern recognition approaches to shape prediction use volumetric representations (e.g.~\cite{rock2015completing,kar2015category}), but more recent works have proposed surface-based representations~\cite{tatarchenko2016multi}. Qi \etal.~\cite{qi2016volumetric} finds an advantage for surface-based representations for 3D object classification, since surfaces can encode high resolution shapes with fewer parameters. Rendered surfaces have fewer pixels than there are voxels in a high resolution mesh.  However, generating a complete shape from 2.5D surfaces creates an additional challenge, since the surfaces need to be aligned and fused into a single 3D object surface.

Second, what is the impact of object-centered vs. view-centered coordinate frames for shape prediction? Nearly all recent 3D shape generation methods use {\em object-centered coordinates}, where the object's shape is represented in a canonical view.  For example, shown either a front view or side view of a car, the goal is to generate the same front-facing 3D model of the car. Object-centered coordinates simplify the prediction problem, but suffer from several practical drawbacks: the viewer-relative pose is not recovered; 3D models used for training must be aligned to a canonical pose; and prediction on novel object categories is difficult due to lack of predefined canonical pose. In {\em viewer-centered coordinates}, the shape is represented in a coordinate system aligned to the viewing perspective of the input image, so a front-view of a car should yield a front-facing 3D model, while a side-view of a car should generate a side-facing 3D model. This increases the variation of predicted models, but also does not require aligned training models and generalizes naturally to novel categories.  

We study these issues using a single encoder-decoder network architecture, swapping the decoder to study volume vs. surface representations and swapping the coordinate frame of predictions to study viewer-centered vs. object-centered.  We examine effects of familiarity by measuring accuracy for novel views of known objects, novel instances of known categories, and objects from novel categories.  We also evaluate prediction from both depth and RGB images. Our experiments indicate a clear advantage for surface-based representations in novel object categories, which likely benefit from the more compact output representations relative to voxels.  Our experiments also show that prediction in viewer-centered coordinates generalizes better to novel objects, while object-centered performs better for novel views of familiar instances. Further, models that learn to predict in object-centered coordinates seem to learn and rely on object categorization to a greater degree than models trained to predict viewer-centered coordinates.

In summary, our main contributions include:
\begin{itemize}
    \item We introduce a new method for surface-based prediction of object shape in a viewer-centered coordinate frame.  Our network learns to predict a set of silhouette and depth maps at several viewpoints relative to the input image, which are then locally registered and merged into a point cloud from which a surface can be computed.
    \item We compare the efficacy of volumetric and surface-based representations for predicting 3D shape, showing an advantage for surface-based representations on unfamiliar object categories regardless of whether final evaluation is volumetric or surface-based.
    \item We examine the impact of prediction in viewer-centered and object-centered coordinates and showing that networks generalize better to novel shapes if they learn to predict in viewer-centered coordinates (which is not currently common practice), and that the coordinate choice significantly changes the embedding learned by the network encoder.
\end{itemize}

\section{Related work}

Our approach relates closely to recent efforts to generate novel views of an object, or its shape. We also touch briefly on related studies in human vision.

\vspace{-3.5mm}
\paragraph{Volumetric shape representations:} 
Several recent studies offer methods to generate volumetric object shapes from one or a few images~\cite{wu20153d, kar2015category, rock2015completing, choy20163d, yan2016prespective, tatarchenko2017octree}. Wu \etal~\cite{wu20153d} proposes a convolutional deep belief network for learning 3D representations using volumetric supervision and evaluate applications to various recognition tasks.  Other studies quantitatively evaluate 3D reconstruction results, with metrics including voxel intersection-over-union~\cite{rock2015completing, choy20163d, yan2016prespective}, mesh distance~\cite{kar2015category, rock2015completing}, and depth map error~\cite{kar2015category}.
Some follow template deformation approaches using surface rigidity~\cite{kar2015category, rock2015completing} and symmetry priors~\cite{rock2015completing}, while others~\cite{wu20153d, choy20163d, yan2016prespective} approach the problem as deep representation learning using encoder-decoder networks.
Fan \etal~\cite{fan2017point} proposes a point cloud generation network that efficiently predicts coarse volumetric object shapes by encoding only the coordinates of points on the surface. Our voxel and multi-surface prediction networks use an encoder-decoder network.  For multi-surface prediction, the decoder generates multiple segmented depth images, pools depth values into a 3D point cloud, and fits a 3D surface to obtain the complete 3D shape.

\vspace{-3.5mm}
\paragraph{Multi-surface representations:} 
Multi-surface representations of 3D shapes are popular for categorization tasks.  The seminal work by Chen \etal~\cite{chen2003visual} proposes a 3D shape descriptor based on the silhouettes rendered from the 20 vertices of a dodecahedron surrounding the object. More recently, Su \etal~\cite{su2015multi} and Qi \etal~\cite{qi2016volumetric} train CNNs on 2D renderings of 3D mesh models for classification.  Qi \etal~\cite{qi2016volumetric} compares CNNs trained on volumetric representations to those trained on multiview representations. Although both representations encode similar amounts of information, they showed that multiview representations significantly outperform volumetric representations for 3D object classification. Unlike our approach, these approaches use multiple projections as input rather than output.

To synthesize multi-surface output representations, we train multiple decoders. Dosovitskiy \etal~\cite{dosovitskiy2015learning} show that CNNs can be used to generate images from high-level descriptions such as object instance, viewpoint, and transformation parameters. Their network jointly predicts an RGB image and its segmentation mask using two up-convolutional output branches sharing a high-dimensional hidden representation. The decoder in our network learns the segmentation for each output view in a similar manner.

Our work is related to recent studies~\cite{tatarchenko2016multi, kulkarni15deep, yang2015weakly, zhu2014multi, yan2016prespective, soltani2017synthesizing, lun20173d} that generate multiview projections of 3D objects. The multiview perceptron by Zhu \etal~\cite{zhu2014multi} generates one random view at a time, given an RGB image and a random vector as input.  Inspired by the mental rotation ability in humans, Yang \etal~\cite{yang2015weakly} proposed a recurrent encoder-decoder network that outputs RGB images rotated by a fixed angle in each time step along a path of rotation, given an image at the beginning of the rotation sequence as input. They disentangle object identity and pose by sharing the identity unit weights across all time steps. Their experiments do not include 3D reconstruction or geometric analysis.

Our proposed method predicts 2.5D surfaces (depth image and object silhouette) of the object from a set of fixed viewpoints evenly spaced over the viewing sphere. In some experiments (Table~\ref{fig:recon_shrec12}), we use 20 views, as in~\cite{chen2003visual}, but we found that 6 views provide similar results and speeds training and evaluation, so 6 views are used for the remainder.  Most existing approaches~\cite{tatarchenko2016multi, kulkarni15deep, yan2016prespective} parameterize the output image as $(x, \theta)$ where $x$ is the input image and $\theta$ is the desired viewpoint relative to canonical object-centered coordinate system. Yan \etal~\cite{yan2016prespective} introduce a formulation that indirectly learns to generate voxels through silhouettes using multi-surface projective constraints, but interestingly they report that voxel IoU performance is better when the network is trained to minimize projection loss alone, compared to when jointly trained with volumetric loss.  

Our approach, in contrast, uses multiview reconstruction techniques (3D surface from point cloud) as a post-process to obtain the complete 3D mesh, treating any inconsistencies in the output images as if they were observational noise. Our formulation also differs in that we learn a view-specific representation, and the complete object shape is produced by simultaneously predicting multiple views of depth maps and silhouettes. In this multi-surface prediction, our approach is similar to Soltani \etal's~\cite{soltani2017synthesizing}, but our system does not use class labels during training. When predicting shape in object-centered coordinates, the predicted views are at fixed orientations compared to the canonical view.  When predicting shape in viewer-centered coordinates, the predicted views are at fixed orientations compared to the input view.

\vspace{-4mm}
\paragraph{Human vision:} 
In experiments on 2D symbols, Tarr and Pinker~\cite{tarr1990does} found that human perception is largely tied to viewer-centered coordinates; this was confirmed by McMullen and Farah~\cite{mcmullen1991viewercentered} for line drawings, who also found that object-centered coordinates seem to play more of a role for familiar exemplars. Note that in the human vision literature, ``viewer-centered'' usually means that the object shape is represented as a set of images in the viewer's coordinate frame, and ``object-centered'' usually means a volumetric shape is represented in the object's coordinate frame.  In our work, we consider both the shape representation (volumetric or surface) and coordinate frame (viewer or object) as separate design choices.  We do not claim our computational approach has any similarity to human visual processing, but it is interesting to see that in our experiments with 3D objects, we also find a preference for object-centered coordinates for familiar exemplars (i.e., novel view of known object) and for viewer-centered coordinates in other cases.

\begin{figure*}
  \begin{center}
  \adjustbox{trim={.022\width} {.00\height} {0.022\width} {.00\height},clip}{
    \includegraphics[scale=0.37]{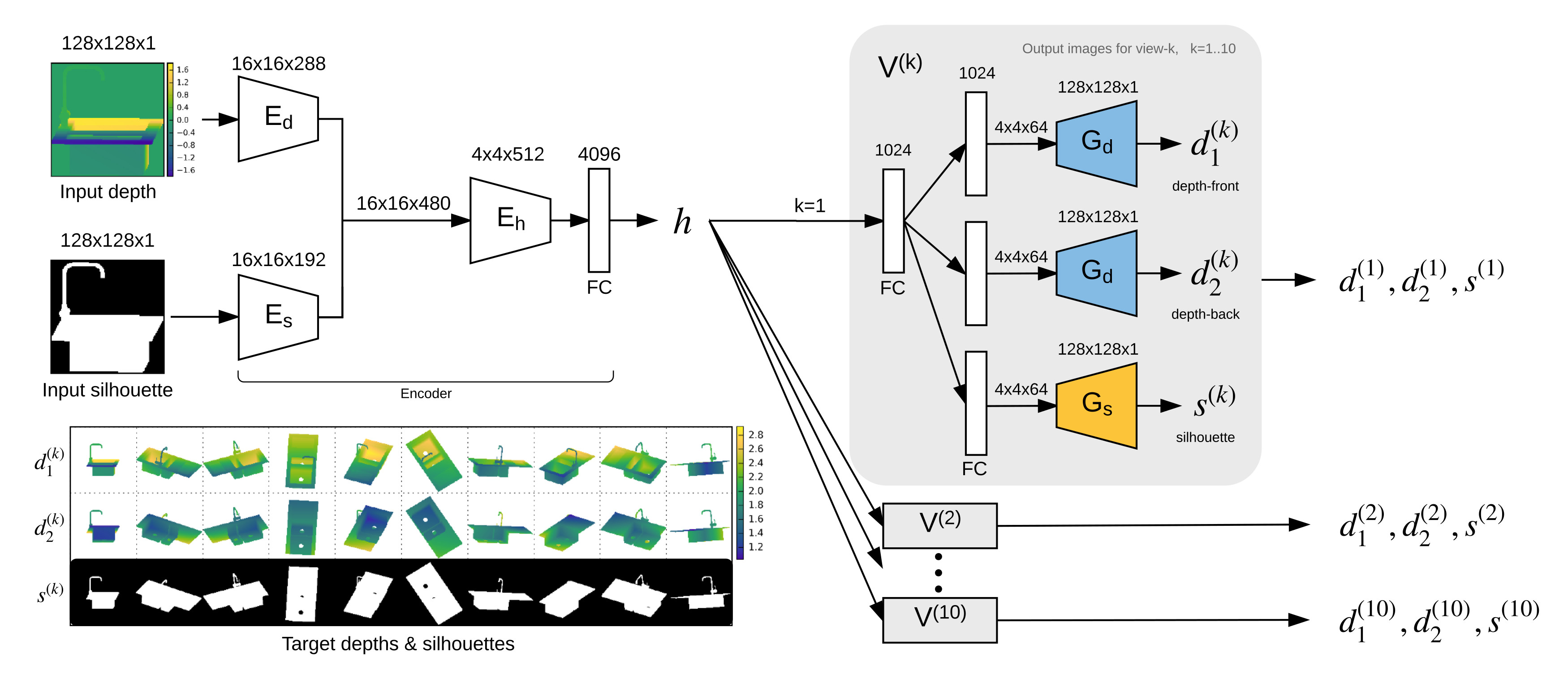}
  }
  \end{center}
  \vspace{-0.1in}
    \caption{Network architecture: Encoders $E_d$, $E_s$, $E_h$ learn view-specific shape features $h$ extracted from the input depth and silhouette. $h$ is used by the 10 output decoder branches $V^{(k)}$, $k$=1..10 which each synthesize one silhouette and two, front and back, depth images. The branches have independently parameterized fully connected layers, but the up-convolutional decoders $G_d$, $G_s$ share parameters across all output branches.
    }
  \label{fig:network}
  \vspace{-0.1in}
\end{figure*}

\section{Viewer-centered 3D shape completion}

Given a single depth or RGB image as input, we want to predict the complete 3D shape of the object being viewed. In the commonly used object-centered setting, the shape is predicted in canonical model coordinates specified by the training data. For example, in the ShapeNetCore dataset, the x-axis or ($\phi_{\text{az}}=\ang{0}, \; \theta_\text{el}= \ang{0}$) direction corresponds to the commonly agreed upon front of the object, and the relative transformation parameters from the input view to this coordinate system is unknown. In our viewer-centered approach, we supervise the network to predict a pre-aligned 3D shape in the input image's reference frame --- e.g. so that $(\phi_{\text{az}}=\ang{0}, \; \theta_\text{el}= \ang{0})$ in the output coordinate system always corresponds to the input viewpoint. Our motivation for exploring these two representations is the hypothesis that networks trained on viewer-centered and object-centered representations learn very different information. A practical advantage of the viewer-centered approach is that the network can be trained in an unsupervised manner across multiple categories without requiring humans to specify intra-category alignment. However, viewer-centered training requires synthesizing separate target outputs for each viewpoint input which increases training data storage cost.

In all experiments, we supervise the networks only using geometric (or photometric) data without providing any side information about the object category label or input viewpoint. The only assumption is that the gravity direction is known (fixed as down in the input view). This allows us to focus on whether the predicted shapes can be completed/interpolated solely based on the 2.5D geometric or RGB input stimuli in a setting where contextual cues are not available. In the case of 2.5D input, we normalize the input depth image so that the bounding box of the silhouette fits inside an orthographic viewing frustum ranging from $\langle {\text{-}1},{\text{-}1} \rangle$ to $\langle 1,1 \rangle$ with the origin placed at the centroid.

\section{Network architectures for shape prediction}

Our multi-surface shape prediction system uses an encoder-decoder network to predict a set of silhouettes and depth maps. Figure~\ref{fig:network} provides an overview of the network architecture, which takes as input a depth map and a silhouette. We also perform experiments on a variant that takes an RGB image as input. To directly evaluate the relative merits of the surface-based and voxel-based representations, we compare this with a volumetric prediction network by replacing the decoder with a voxel generator.
Both network architectures can be trained to produce either viewer-centered or object-centered predictions.

\subsection{Generating multi-surface depth and silhouettes}

We observe that, for the purpose of 3d reconstruction, it is important to be able to see the object from certain viewpoints -- e.g. classes such as cup and bathtub need at least one view from the top to cover the concavity.
Our proposed method therefore predicts 3D object shapes at evenly spaced views around the object. We place the cameras at the 20 vertices $\{\mathbf{v}_0, .., \mathbf{v}_{19}\}$ of a dodecahedron centered at the origin. A similar setup was used in the Light Field Descriptor \cite{chen2003visual} and a recent study by Soltani \etal \cite{soltani2017synthesizing}.

In order to determine the camera parameters, we rotate the vertices so that vertex $\mathbf{v}_0=\langle 1,1,1 \rangle$ aligns with the input viewpoint in the object's model coordinates. The up-vectors point towards the z-axis and are rotated accordingly.
Note that the input viewpoint $\mathbf{v}_0$ is not known in our setting, but the relative transformations from $\mathbf{v}_0$ to all of the output viewpoints are known and fixed.

As illustrated in Figure~\ref{fig:network}, our network takes the depth image and the silhouette in separate input branches. The encoder units ($E_d$, $E_s$, $E_h$) consist of bottleneck residual layers. $E_d$ and $E_s$ each take in a depth image and a silhouette. They are concatenated in the channel dimension at resolution 16 and the following residual layers $E_h$ output the latent vector $h$ from which all output images are derived simultaneously.
An alternate approach is taking in a two-channel image in a single encoder. We experimented with both architectures and found the two-branch network to perform better.

We use two generic decoders (Table~\ref{fig:network}) to generate the views, one for all depths and another for all silhouettes.  Each view in our setting has a corresponding segmented silhouette and another view on the opposite side, thus only 10 out of the 20 silhouettes need to be predicted due to symmetry (or 3 out of 6 if predicting six views).
The network therefore outputs a silhouette and corresponding front and back depth images $\{s^{(i), d_f^{(i)}, d_b^{(i)}} \}$ in the $i$-th output branch.
Similarly to Dosovitskiy \etal~\cite{dosovitskiy2015learning}, we minimize the objective function 

\vspace{-3mm}
$$\mathbf{L}_{\text{proj}} = k \mathbf{L}_s + (1-k) \mathbf{L}_d $$

where $\mathbf{L}_s$ is the mean logistic loss over the silhouettes and $\mathbf{L}_d$ is the mean MSE over the depth maps whose silhouette label is 1. We use $k=0.2$ in our experiments.

\begin{figure}[t]
\begin{center}
  \includegraphics[width=0.8\linewidth]{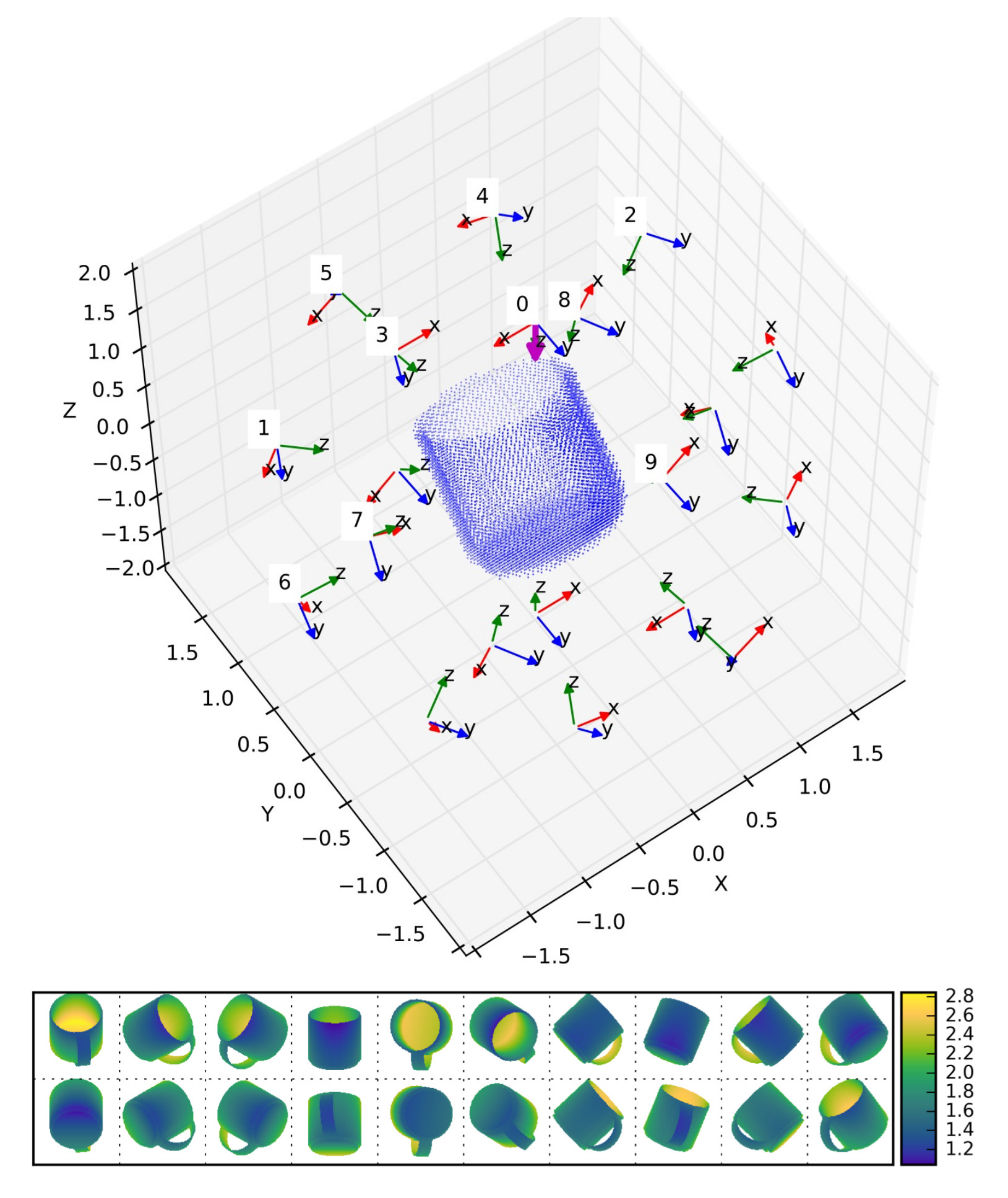}
\end{center}
\vspace{-0.1in}
  \caption{Illustration of ground truth generation. Top: GT mesh voxelized in camera coordinates which are used in the viewer-centered voxel baseline experiment. The numeric labels around the model are the indices of the viewpoints from which the multi-surface depth images (shown above) were rendered, left to right. Viewpoint-0, whose camera is located at the origin, \textsl{always} corresponds to the input view, and the relative transformation from Viewpoint-0 to all the other viewpoints is constant throughout all our experiments.
  Bottom: Multi-surface projections are offset relative the the input view. This viewer-centered approach does not require alignment between the 3D models and allows the network to be trained in an unsupervised manner on synthetic data.
  }
  \vspace{-0.1in}
\label{fig:datasetPrep}
\label{fig:onecol}
\end{figure}

\subsection{Reconstructing multi-surface representations} 

In our study, we use the term ``reconstruction'' to refer to surface mesh reconstruction in the final post-processing step.  We convert the predicted multiview depth images to a single triangulated mesh using Floating Scale Surface Reconstruction (FSSR) \cite{fuhrmann2014floating}, which we found to produce better results than Poisson Reconstruction \cite{kazhdan2013screened} in our experiments. FSSR is widely used for surface reconstruction from oriented 3D points derived from multiview stereo or depth sensors.
Our experiments are unique in that surface reconstruction methods are used to resolve noise in predictions generated by neural networks rather than sensor observations.  We have found that 3D surface reconstruction reduces noise and error in surface distance measurements.

\subsection{Generating voxel representations}

We compare our multi-surface shape prediction method with a baseline that directly predicts a 3D voxel grid.
Given a single-view depth image of an object, the ``Voxels'' network generates a grid of 3D occupancy 
mappings in the camera coordinates of viewpoint $\mathbf{v}_0$. The cubic window of length 2 centered at $\langle 0,0,1 \rangle$ is voxelized after camera transformation. 
The encoded features $h$ feed into 3D up-convolutional layers, outputting a final $48 \times 48 \times 48$ volumetric grid. The network is trained from scratch to minimize the logistic loss $\mathbf{L}_v$ over the binary voxel occupancy labels.

\section{Experiments}

We first describe the datasets (Sec.~\ref{sec:exp_dataset}) and evaluation metrics (Sec.~\ref{sec:exp_metrics}), then discuss results in Section~\ref{sec:discussion}. 
In all experiments, we train the networks on synthetically generated images.  
A single training example for the multi-surface network is the input-output pair $(x_d, x_s) \rightarrow \{(s^{(k)}, d_f^{(k)}, d_b^{(k)}) \}_{k= 0..9}$ where $(x_d, x_s)$ is the input depth image and segmentation, and the orthographic depth images $(s^{(k)}, d_f^{(k)}, d_b^{(k)})$ serve as the output ground truth. The $k$-th ground truth silhouette has associated front and back depth images. Each image is uniformly scaled to fit within 128x128 pixels.
Training examples for the voxel prediction network consist of input-output pairs $(x_d, x_s) \rightarrow V$, where $V$ is a grid of ground truth voxels (size 48x48x48 for the input depth experiments, and 32x32x32 for the input RGB experiments).

\subsection{Datasets}
\label{sec:exp_dataset}

\begin{figure*}
  \begin{center}
  \adjustbox{trim={.00\width} {.003\height} {0.00\width} {.01\height},clip}{
    \includegraphics[scale=0.31]{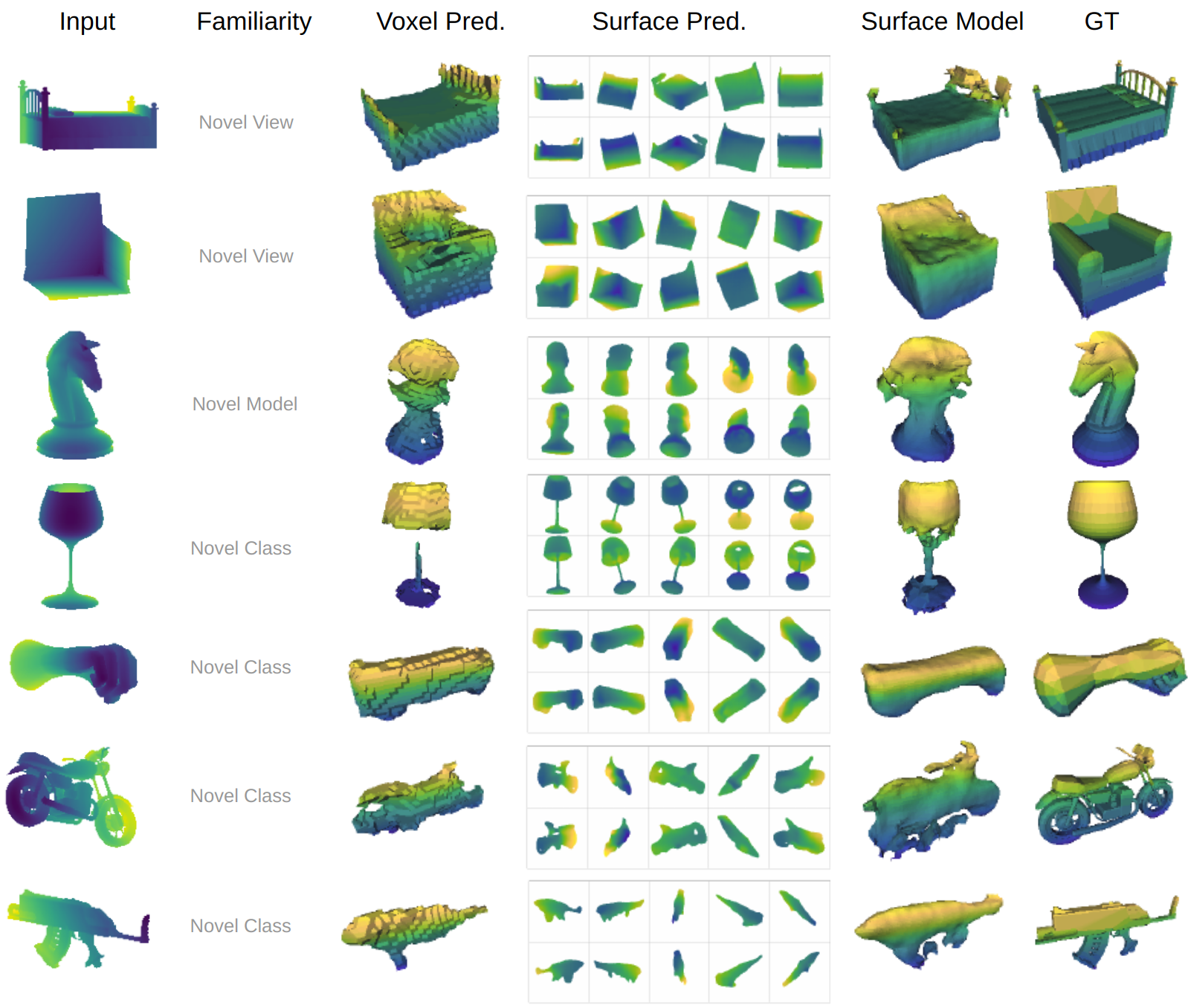}
  }
  \end{center}
  \vspace{-0.1in}
    \caption{Multi-surfaces vs. Voxels. ``Novel View'' means other views of that shape instance were seen during training.  ``Novel Model'' means other instances from the same class were seen in training.  ``Novel Class'' means that no instances from that category were seen during training.  Under multi-surface, a subset of the predicted depth maps are shown, as well as the complete reconstructed shape. The multi-surface approach tends to generalize better for novel classes.}
  \label{fig:qual_depth}
\end{figure*}

\begin{table*}
\begin{center}
    \normalsize
\begin{tabular}{ | l || c|c|c  || c|c|c |}
\hline
    {{Mean }}  &  \multicolumn{3}{ c ||}{Surface Distance} &  \multicolumn{3}{ c |}{Voxel IoU} \\
    \cline{2-7}
    {} & NovelClass &NovelModel  &NovelView  & NovelClass &NovelModel  &NovelView  \\ 
\hline
    Voxels      & 0.0950           & 0.0619           & 0.0512           & 0.4569           & 0.5176           & \bfseries 0.6969 \\
    Multi-surfaces  & \bfseries 0.0759 & 0.0622           & \bfseries 0.0494 & 0.4914           & 0.5244           & 0.6501           \\
    Rock \etal \cite{rock2015completing} & 0.0827           & \bfseries 0.0604 & 0.0639           & \bfseries 0.5320 & \bfseries 0.5888 & 0.6374           \\

\hline
\end{tabular}

\end{center}
\caption{
  3D shape prediction from a single depth image on the SHREC'12 dataset used by~\cite{rock2015completing}, comparing results for voxel and multi-surface decoders trained to produce models in a viewer-centered coordinate frame.  Rock \etal.~\cite{rock2015completing} also predicts in viewer-centered coordinates.
}
\label{fig:recon_shrec12}
\end{table*}

\begin{table}[]
\small
\begin{center}
\begin{tabular}{|l|c|c|c|}
\hline
                   & \small{NovelView} & \small{NovelModel}  & \small{NovelClass}\\
\hline
View-centered     &    0.714  &  \bf{0.570}      &  \bf{0.517}      \\
Obj-centered     &    \bf{0.902}  &  0.474      &  0.309      \\
\hline
\end{tabular}
\end{center}
\caption{
Voxel IoU of predicted and ground truth values (mean, higher is better), using
the voxel network. Trained for 45 epochs with batch size 150, learning rate 0.0001.
}
\label{table:shrec12_objview_iou}
\end{table}

\begin{table}[]
\small
\begin{center}
\begin{tabular}{|l|c|c|c|}
\hline
                   & \small{NovelView} & \small{NovelModel}  & \small{NovelClass}\\
\hline
View-centered     &    0.807  &  \bf{0.706}      &  \bf{0.670}      \\
Obj-centered     &    \bf{0.921}  &  0.586 &  0.416      \\
\hline
\end{tabular}
\end{center}
\caption{
Silhouette IoU, using the 6-view multi-surface network (mean, higher is better).
}
\label{table:shrec12_objview_silh}
\end{table}

\begin{table}[]
\small
\begin{center}
\begin{tabular}{|l|c|c|c|}
\hline
                   & \small{NovelView} & \small{NovelModel}  & \small{NovelClass}\\
\hline
View-centered     &    0.011  &  \bf{0.016}      &  \bf{0.0207}      \\
Obj-centered     &    \bf{0.004}  & 0.035  &  0.0503 \\
\hline
\end{tabular}
\end{center}
\caption{
Depth error, using the 6-view multi-surface network (mean, lower is better).
}
\label{table:shrec12_objview_depth}
\vspace{-2.4mm}
\end{table}

\boldhead{3D shape from single depth} We use the SHREC'12 dataset for comparison with the exemplar retrieval approach by Rock \etal~\cite{rock2015completing} on predicting novel views, instances, and classes.  Novel views require the least generalization (the same shape is seen in training), and novel classes require the most (no instances from the same category seen during training). This dataset has a training set consisting of 22,500 training + 6,000 validation examples and has 600 examples in each of the three test evaluation sets,
using the standard splits~\cite{rock2015completing}.
The 3D models in the dataset are aligned to each other, so that they can be used for both viewer-centered and object-centered prediction.  Results are shown in Fig.~\ref{fig:qual_depth} and Tables~\ref{fig:recon_shrec12}, \ref{table:shrec12_objview_iou}, \ref{table:shrec12_objview_silh}, and \ref{table:shrec12_objview_depth}. 

\begin{figure*}
  \begin{center}
  \adjustbox{trim={.00\width} {.00\height} {0.00\width} {.00\height},clip}{
    \includegraphics[scale=0.26]{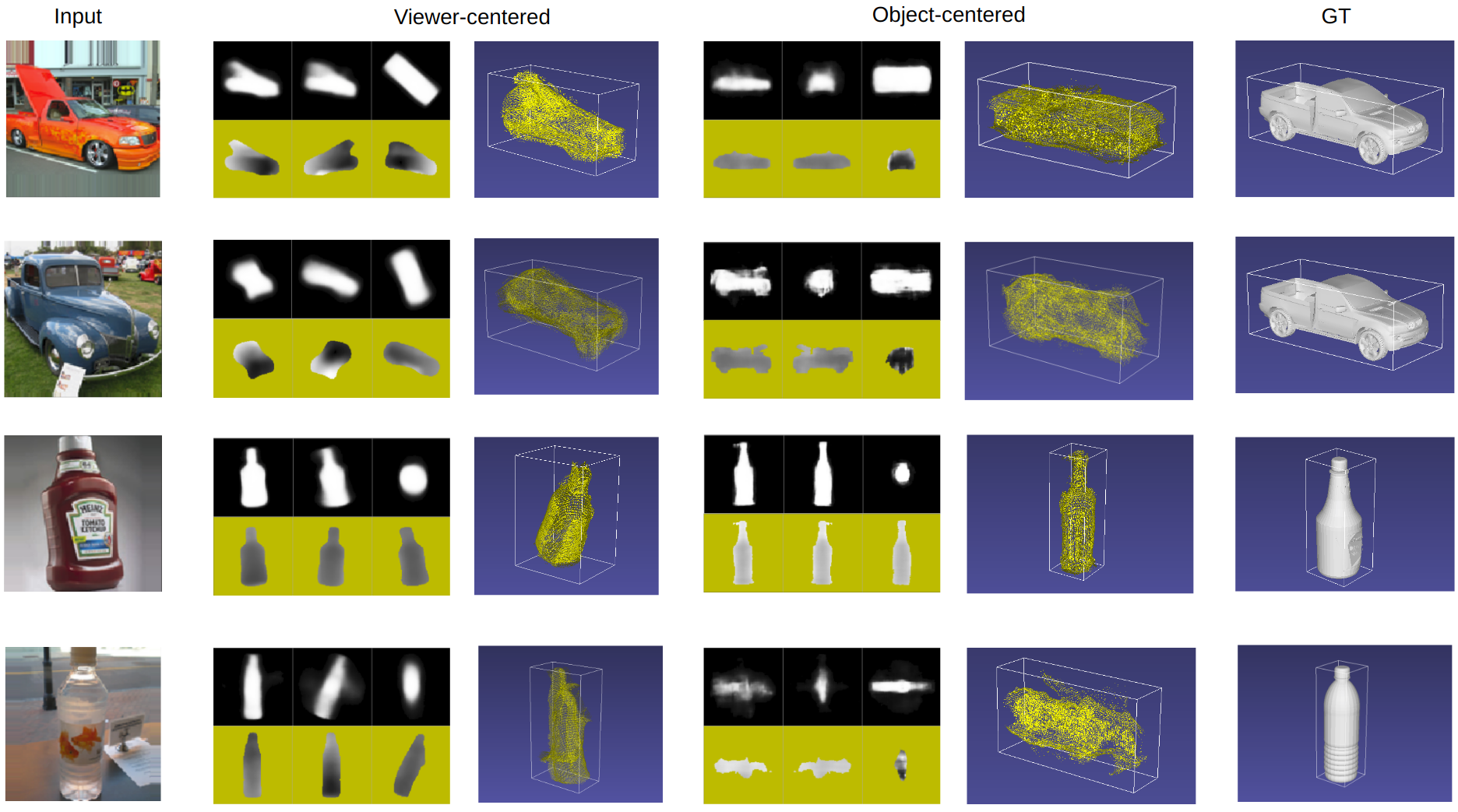}
  }
  \end{center}
  \vspace{-0.1in}
    \caption{\textbf{RGB-based shape prediction examples: } On left, is the input image.  We show predicted depth maps and silhouettes from three views and a merged point cloud from all views, produced by the networks trained with object-centered coordinates and with viewer-centered coordinates.  Viewer-centered tends to generalize better while object-centered sometimes produces a model that looks good but is from entirely the wrong category.  In viewer-centered, the encoder learns to map inputs together if they correspond to similar shapes in similar poses, learning a viewpoint-sensitive representation.  In object-centered, the encoder learns to map different views of the same object together, learning a viewpoint-invariant representation.  
    }
    \vspace{-0.1in}
  \label{fig:qual_rgb}
\end{figure*}

\begin{table}[]
    \footnotesize
\begin{center}
\begin{tabular}{|l|c|c|c|c|}
\hline
Category & \cite{choy20163d} (OC) & \cite{choy20163d} (VC) & Ours (OC) & Ours (VC) \\
\hline
aero          &   0.359    & 0.201   &  \textbf{0.362}        & 0.289  \\
bike          &   \textbf{0.535}    & 0.106   &  0.362        & 0.272  \\
boat          &   \textbf{0.366}    & 0.236   &  0.331        & 0.259  \\
bottle        &   0.617    & 0.454   &  \textbf{0.643}        & 0.576  \\
bus           &   0.387    & 0.273   &  0.497        & \textbf{0.556}  \\
car           &   0.462    & 0.400   &  0.566        & \textbf{0.582}  \\
chair         &   0.325    & 0.221   &  \textbf{0.362}        & 0.332  \\
d.table       &   0.081    & 0.023   &  \textbf{0.122}        & 0.118  \\
mbike         &   0.474    & 0.167   &  \textbf{0.487}        & 0.366  \\
sofa          &   \textbf{0.602}    & 0.447   &  0.555        & 0.538  \\
train         &   \textbf{0.340}    & 0.192   &  0.301        & 0.257  \\
tv            &   0.376    & 0.164   &  0.383        & \textbf{0.397}   \\
\hline                      
mean          &   0.410     & 0.240  &  \textbf{0.414}        & 0.379  \\
\hline
\end{tabular}
\end{center}
\caption{
Per-category voxel IoU on PASCAL 3D+ using our multi-surface network and the voxel-based 3D-R2N2 network~\cite{choy20163d}.  Although the network trained to produce object-centered (OC) models performs slightly better quantitatively (for multi-surface), the viewer-centered (VC) model tends to produce better qualitative results (see supplemental material for more), sometimes with misaligned pose.  
}
\label{table:pascal3d}
\vspace{-3mm}
\end{table}

\vspace{0.3mm}
\boldhead{3D shape from real-world RGB images}
We also perform novel model experiments on RGB images. We use RenderForCNN's \cite{su2015render} rendering pipeline and generate 2.4M synthetic training examples using the ShapeNetCore dataset along with target depth and voxel representations. In this dataset, there are 34,000 3D CAD models from 12 object categories. 
We perform quantitative evaluation of the resulting models on real-world RGB images using the PASCAL 3D+ dataset~\cite{xiang_wacv14_pascal3d}. We train 3D-R2N2's network~\cite{choy20163d} from scratch using the same dataset and compare evaluation results. The results we report here differ from those in the original paper due to differences in the training and evaluation sets.  Specifically, the results reported in~\cite{choy20163d} are obtained after fine-tuning on the PASCAL 3D+ dataset, which is explicitly discouraged in~\cite{xiang_wacv14_pascal3d} because the same 3D model exemplars are used for train and test examples.

Thus, we train on renderings and test on real RGB images of objects that may be partly occluded and have background clutter -- a  challenging task.  Results are shown in Tables~\ref{table:pascal3d} and in Figure~\ref{fig:qual_rgb}.

\subsection{Evaluation metrics and processes}
\label{sec:exp_metrics}

\boldhead{Voxel intersection-over-union}
Given a mesh reconstructed from the multi-surface prediction (which may not be watertight), we obtain a solid representation by voxelizing the mesh surface into a hollow volume and then filling in the holes using ray tracing. All voxels not visible from the outside are filled. Visibility is determined as follows: from the center of each voxel, we scatter 1000 rays and the voxel is considered visible if any of them can reach the edge of the voxel grid. We compute intersection-over-union (IoU) with the corresponding ground truth voxels, defined as the number of voxels filled in both representations divided by the number of voxels filled in at least one.

\boldhead{Surface distance}
We also evaluate with a surface distance metric similar to \cite{rock2015completing}, which tends to correspond better to qualitative judgments of accuracy when there are thin structures.
The distance between surfaces is approximated as the mean of \textsl{point-to-triangle} distances from i.i.d. sampled points on the ground truth mesh to the closest points on surface of the reconstructed mesh, and vice versa. We utilize a KD-tree to find the closest point on the mesh.  To ensure scale invariance of this measure across datasets, we divide the resulting value by the mean distance between points sampled on the GT surface. The points were sampled at a density of 300 points per unit area.
To evaluate surface distance for voxel-prediction models, we use Marching Cubes to obtain the mesh from the prediction.

\boldhead{Image-based measures} For multi-surface experiments, in addition to voxel IoU and surface distance, we also evaluate using silhouette intersection-over-union and depth error averaged over the predicted views. Sometimes, even when the predictions for individual views are quite accurate, slight inconsistencies or oversmoothing by the final surface estimation can reduce the accuracy of the 3D model.

\section{Discussion}
\label{sec:discussion}

\paragraph{Multi-surface vs. voxel shape representations:}
Table~\ref{fig:recon_shrec12} compares performance of multi-surface and voxel-based representations for shape prediction.
Quantitatively, multi-surface outperforms for novel class and performs similarly for novel view and novel instance.
We also find that the 3D shapes produced by the multi-surface model look better qualitatively, as they can encode higher resolution.

We observe that it is generally difficult to learn and reconstruct thin structures such as the legs of chairs and tables. In part this is a learning problem, as discussed in Choy \etal.~\cite{choy20163d}. Our qualitative results suggest that silhouettes are generally better for learning and predicting thin object parts than voxels, but the information is often lost during surface reconstruction due to the sparsity of available data points. We expect that improved depth fusion and mesh reconstruction would likely yield even better results.
As shown in Fig. 6, the multi-surface representation can more directly be output as a point cloud by skipping the reconstruction step. This avoids errors that can occur during the surface reconstruction but is more difficult to quantitatively evaluate. 

\vspace{0.4mm}
\boldhead{Viewer-centered vs. object-centered coordinates}
When comparing performance of predicting in viewer-centered coordinates vs. object-centered coordinates, it is important to remember that only viewer-centered encodes pose and, thus, is more difficult. Sometimes, the 3D shape produced by viewer-centered prediction is very good, but the pose is misaligned, resulting in poor quantitative results for that example.  Even so, in Tables~\ref{table:shrec12_objview_iou}, \ref{table:shrec12_objview_silh}, and \ref{table:shrec12_objview_depth}, we observe a clear advantage for viewer-centered prediction for novel models and novel classes, while object-centered outperforms for novel views of object instances seen during training. For object-centered prediction, two views of the same object should produce the same 3D shape, which encourages memorizing the observed meshes.  Under viewer-centered, the predicted mesh must be oriented according to the input viewpoint, so multiple views of the same object should produce different 3D shapes (which are related by a 3D rotation).  This requirement seems to improve the generalization capability of viewer-centered prediction to shapes not seen during training. 

In Table~\ref{table:pascal3d}, we see that object-centered prediction quantitatively slightly outperforms for RGB images. In this case, training is performed on rendered meshes, while testing is performed on real images of novel object instances from familiar categories. Qualitatively, we find that viewer-centered prediction tends to produce much more accurate shapes, but that the pose is sometimes wrong by 15-20 degrees, perhaps as a result of dataset transfer.  

Qualitative results support our initial hypothesis that object-centered models tend to correspond more directly to category recognition.  We see in Figure~\ref{fig:qual_rgb}, that the object-centered model often predicts a shape that looks good but is an entirely different object category than the input image.  The viewer-centered model tends not to make these kinds of mistakes and, instead, errors tend to be overly simplified shapes or slightly incorrect poses.

\vspace{0.4mm}
\boldhead{Implications for object recognition}
While not a focus of our study, we also trained an object classifier using the 4096-dimensional encoding layer of the viewer-centric model as input features for a single hidden-layer classifier. The resulting classifier outperformed by $1\%$ a Resnet classifier that was trained end-to-end on the same data. This indicates that models trained to predict shape and pose contain discriminative information that is highly useful for predicting object categories and may, in some ways, generalize better than models learned to predict categories directly. More study is needed in this direction.

\section{Conclusion}

Recent methods to produce 3D shape from a single image have used a variety of representation for shape (voxels, octrees, multiple depth maps).  By utilizing the same encoder
architecture for volumetric and surface-based representations, we are able to more directly compare their efficacy.  Our experiments
show an advantage for surface-based representations in predicting novel object shapes, likely because they can encode shape details with fewer parameters. Nearly all existing methods predict object shape in object-centered
coordinates, but our experiments show that learning to predict shape in viewer-centered coordinates leads to better generalization for novel objects.  Further improvements in surface-based
prediction could be obtained through better alignment and fusing of produced depth maps. More research is also needed to verify whether recently proposed octree-based representations~\cite{tatarchenko2017octree} close the gap with surface-based representations.  In addition, the relationship between object categorization and shape/pose prediction requires
further exploration. Novel view prediction and shape completion could provide a basis for unsupervised learning of features that are effective for object category and attribute recognition.

\vspace{0.74mm}
\boldhead{Acknowledgements}
This project is supported by NSF Awards IIS-1618806, IIS-1421521, Office of Naval Research grant ONR MURI N00014-16-1-2007 and a hardware donation from NVIDIA.

{\small
\bibliographystyle{ieee}
\bibliography{mvshape_bib}
}

\end{document}